\pdfoutput=1

\documentclass[11pt]{article}

\usepackage{acl}

\usepackage[utf8]{inputenc} 
\usepackage[T1]{fontenc}    
\usepackage{hyperref}       
\usepackage{url}            
\usepackage{booktabs}       
\usepackage{amsfonts}       
\usepackage{nicefrac}       
\usepackage{microtype}      
\usepackage{xcolor}         

\usepackage{times}
\usepackage{latexsym}
\usepackage{amsmath}
\usepackage{pifont}
\usepackage{graphicx}
\usepackage{listings}

\usepackage{subcaption}

\renewcommand{\O}[1]{$\mathcal{O}(#1)$}

\title{FNet: Mixing Tokens with Fourier Transforms}

\author{
  James Lee-Thorp \and Joshua Ainslie \and Ilya Eckstein \and Santiago Onta\~{n}\'{o}n \\
  Google Research \\
  \texttt{\{jamesleethorp, jainslie, ilyaeck, santiontanon\}@google.com}
}

\begin{document}

\maketitle

\begin{abstract}
  We show that Transformer encoder architectures can be sped up, with limited accuracy costs, by replacing the self-attention sublayers with simple linear transformations that ``mix'' input tokens. Most surprisingly, we find that replacing the self-attention sublayer in a Transformer encoder with a standard, unparameterized Fourier Transform achieves $92$-$97\%$ of the accuracy of BERT counterparts on the GLUE benchmark, but trains $80\%$ faster on GPUs and $70\%$ faster on TPUs at standard 512 input lengths. At longer input lengths, our FNet model is significantly faster:  when compared to the ``efficient Transformers'' on the Long Range Arena benchmark, FNet matches the accuracy of the most accurate models, while outpacing the fastest models across all sequence lengths on GPUs (and across relatively shorter lengths on TPUs). Finally, FNet has a light memory footprint and is particularly efficient at smaller model sizes; for a fixed speed and accuracy budget, small FNet models outperform Transformer counterparts.\footnote{Code is available at \url{https://github.com/google-research/google-research/tree/master/f_net}.}
\end{abstract}

\section{Introduction}
\label{sec:introduction}

The Transformer architecture \citep{vaswani2017attention} has achieved rapid and widespread dominance in NLP. At its heart is a attention mechanism -- an inductive bias that connects each token in the input through a relevance weighted basis of every other token. Many papers have prodded and probed the Transformer, and in particular the attention sublayers, in an effort to better understand the architecture; see, for example, \citet{tenney-etal-2019-bert, vig-belinkov-2019-analyzing, clark-etal-2019-bert, voita-etal-2019-analyzing}. Although potentially limited in their effectiveness \citep{hewitt-liang-2019-designing}, these probes generally back the intuition that, by allowing higher order units to form out of compositions of the input, Transformer models can flexibly capture diverse syntactic and semantic relationships. 

In this work, we investigate whether simpler token mixing mechanisms can wholly replace the relatively complex self-attention layers in Transformer encoder architectures. We first replace the attention sublayer with two parameterized matrix multiplications -- one mixing the sequence dimension and one mixing the hidden dimension. Seeing promising results in this simple linear mixing scheme, we further investigate the efficacy of faster, structured linear transformations. Surprisingly, we find that the Fourier Transform, despite having no parameters at all, achieves nearly the same performance as dense linear mixing and scales very efficiently to long inputs, especially on GPUs (owing to the \O{N \log N} Fast Fourier Transform (FFT) algorithm). We call the resulting model FNet. 

While Fourier Transforms have previously been used to approximate or speed up computations in Convolutional Neural Networks~\citep{el2004fast, mathieu2014fast, highlander2016very, pratt2017fcnn, lin2018fft, chitsaz2020acceleration, goldberg2020rethinking}, Recurrent Neural Networks \citep{koplon1997using, zhang2000forenet, zhang2018learning},  Transformers~\citep{choromanski2020masked, tamkin2020language}, and MLP layers more generally~\citep{cheng2015exploration, moczulski2015acdc, sindhwani2015structured}, we believe our work is the first to wholly replace particular neural network sublayers with a Fourier Transform. This approach of viewing the Fourier Transform as a first class mixing mechanism is reminiscent of the MLP-Mixer~\citep{tolstikhin2021mlp} for vision, which replaces attention with MLPs; although in contrast to MLP-Mixer, FNet has no learnable parameters that mix along the spatial dimension.

Given the favorable asymptotic complexity of the FFT, our work also connects with the literature on ``long sequence'' or ``efficient'' Transformers, which aim to make the attention mechanism scale better via sparsity patterns~\citep{child2019generating, qiu2019blockwise, parmar2018image, beltagy2020longformer, ainslie2020etc, zaheer2020big, wang2020linformer, tay2020sparse, tay2020synthesizer, kitaev2020reformer, roy2021efficient, vyas2020fast, liu2018generating} or via linearization of the attention matrix~\citep{katharopoulos2020transformers, choromanski2020rethinking, peng2021random}. As we will show in our experiments, while some of those works achieve $O(N)$ scaling of attention, this complexity often hides large constants, which make them less scalable in practice than FNet.

The contributions of our paper are:
\begin{itemize}
    \item We show that simple linear transformations, including even (parameter-free) Fourier Transforms, along with standard MLPs in feed-forward layers, are competent at modeling diverse relationships in text. That such a simple linear transformation works at all is surprising, and suggests that, for at least some NLP problems, attention may not be the principal component driving the performance of Transformers.
    \item We introduce a new model, FNet, that uses the Fourier Transform as a mixing mechanism.  FNet offers an excellent compromise between speed, memory footprint, and accuracy, achieving $92\%$ and $97\%$, respectively, of the accuracy of BERT-Base and BERT-Large \citep{devlin2018bert} on the GLUE benchmark \citep{wang2018glue}, while training $80\%$ faster on GPUs and $70\%$ faster on TPUs. 
    \item We find that FNet hybrid models containing only two self-attention sublayers achieve $97-99\%$ of their BERT counterparts' accuracy on GLUE, while still running $40-70\%$ faster. This indicates that, while attention can improve accuracy, it may not be necessary to use in every layer.
    \item We demonstrate FNet scales very well to long inputs and offers a better compromise between speed and accuracy than the efficient Transformers evaluated on the Long-Range Arena (LRA) benchmark \citep{tay2020long}.
    Specifically, FNet achieves accuracy comparable to the most accurate efficient Transformer architectures but is significantly faster at both training and inference than all of the evaluated Transformer architectures across all sequence lengths on GPUs. On TPUs, FNet is faster for relatively shorter sequence lengths; for longer sequences, the only efficient Transformers that are faster than FNet on TPUs are less accurate on the LRA benchmark.
    Based on this, we argue that rather than seeking more efficient approximations of the attention, there may be more value in seeking out completely new mixing mechanisms.
\end{itemize}

\section{Related work}
\label{sec:related_work}

\subsection{Fourier Transforms in neural networks}
\label{subsec:fourier_transforms_neural_networks}

Fourier analysis features heavily in studies of the universal approximation properties of neural networks; see, for example, \citep{cybenko1989approximation, barron1993universal}. In terms of practical applications, discrete Fourier Transforms (DFT), and in particular the Fast Fourier Transform (FFT), have been used to tackle signal processing problems such as fitting neural networks to FFTs of electrocardiogram signals \citep{minami1999real, gothwal2011cardiac, mironovova2015fast} and vibration signals \citep{zhang2013fault}, or to evolve solutions of Partial Differential Equations \citep{li2020fourier}.

Because ordinary multiplication in the frequency domain corresponds to a convolution in the time domain, FFTs have been deployed in Convolutional Neural Networks to speed up computations, in Recurrent Neural Networks to speed up training and reduce exploding and vanishing gradients, and generally to approximate dense, linear layers to reduce computational complexity; see references cited in Section \ref{sec:introduction}. DFTs have also been used indirectly in several Transformer works. The Performer \citep{choromanski2020masked} linearizes the Transformer self-attention mechanism by leveraging random Fourier features to approximate a Gaussian representation of the softmax kernel. In our work, rather than approximating attention, we replace attention with the Fourier Transform, which acts as an alternate hidden representation mixing mechanism.  \citet{tamkin2020language} use spectral filters to generate hierarchical features, showing that the filtered embeddings perform well in different tasks (word-level, sentence-level or document-level), depending on which frequency scales are filtered. In contrast to FNet, they separate Fourier frequencies, rather than using the transform to combine features.
Finally, through personal communication, we were alerted to concurrent, unpublished work \citep{backurs2021note} that describes an FFT based neural model that is very similar to FNet.

\subsection{Modeling semantic relations via attention}
\label{subsec:power_of_attention}

Attention models have achieved state of the art results across virtually all NLP tasks and even some image tasks \citep{dosovitskiy2020image}.
This success is generally attributed to the flexibility and capacity of attention. Although some works \citep{ramsauer2020hopfield} have endeavoured to gain a deeper understanding of attention, the pervasive intuition is that the success of attention models derives from the token-dependent attention patterns in different layers; see, for example, \citep{tenney-etal-2019-bert}. However, it is natural to ask: Do we really need the flexibility, and associated cost, of attention? 

\citet{tay2020synthesizer} empirically investigated the importance of the dot product operation in the attention mechanism in their Synthesizer model (related to our ``Linear'' baseline below). They find that learnt token-dependent attention weights are highly expressive, but not necessarily crucial for realizing accurate NLP models. \citet{you2020hard} replace attention weights in the Transformer encoder and decoder with unparameterized Gaussian distributions, showing minimal performance degradation provided they retain learnable cross-attention weights. Similarly, \citet{raganato2020fixed} find little to no accuracy degradation when replacing all but one of the attention heads of each attention layer in the \emph{encoder} with fixed, non-learnable positional patterns. Finally, \citet{tolstikhin2021mlp} present MLP-Mixer, where attention is replaced by MLPs, with limited performance degradation in image classification tasks.

\subsection{Efficient and long sequence models}
\label{subsec:efficient_transformers}

The standard attention mechanism \citep{vaswani2017attention} has a quadratic time and memory bottleneck with respect to sequence length. This limits its applicability in tasks involving long range dependencies. Most efforts to improve attention efficiency are based on sparsifying the attention matrix. \citet{tay2020efficient} survey many of the recent efficient attention works; see also citations in Section \ref{sec:introduction}. Several ``efficient Transformers'' achieve \O{N\sqrt{N}} or even \O{N} theoretical complexity. However, the constants hidden by this notation can be large. For example, in models such as Longformer \citep{beltagy2020longformer}, ETC \citep{ainslie2020etc}, and BigBird \citep{zaheer2020big}, attention is \O{N} as a function of the input length, but quadratic in the number of ``global tokens''; the latter must be sufficiently large to ensure good performance.

The Long-Range Arena benchmark \citep{tay2020long} attempts to compare many of the efficient Transformers in a series of tasks requiring long range dependencies, finding that the Performer \citep{choromanski2020rethinking}, Linear Transformer \citep{katharopoulos2020transformers}, Linformer \citep{wang2020linformer}, and Image Transformer (Local Attention) \citep{parmar2018image} were the fastest on TPUs and had the lowest peak memory usages per device.\footnote{Memory usage is often overlooked, but empirical studies have shown that Transformer architectures are often memory-bound \citep{ivanov2020data, shazeer2019fast}.} Instead, in this paper we completely replace self-attention with a different mixing, namely the Fourier Transform, which offers: (1) performance,
(2) reduced model size (no learnable parameters), and (3) simplicity.

Finally, we note that, in an effort to investigate different token mixing mechanisms, we compare a vanilla BERT model \citep{devlin2018bert} with a vanilla FNet, ignoring more recent Transformer optimizations, which we consider orthogonal to this work; see, for example, \citep{narang2021transformer, kim2020fastformers, shleifer2020pre}.

\section{Model}
\label{sec:model}

\subsection{Discrete Fourier Transform}
\label{subsec:dft_background}

The Fourier Transform decomposes a function into its constituent frequencies. Given a sequence $\{x_n\}$ with $n \in [0, N-1]$, the discrete Fourier Transform (DFT) is defined by the formula:
\begin{equation}
    \label{eq:dft}
    X_k = \sum_{n=0}^{N-1} x_n e^{-{\frac{2\pi i}{N}} n k}, \quad 0 \leq k \leq N-1.
\end{equation}
For each $k$, the DFT generates a new representation $X_k$ as a sum of all of the original input tokens $x_n$, with so-called ``twiddle factors''.
There are two primary approaches to computing the DFT: the Fast Fourier Transform (FFT) and matrix multiplication. The standard FFT algorithm is the Cooley–Tukey algorithm \citep{cooley1965algorithm, frigo2005design}, which recursively re-expresses the DFT of a sequence of length $N = N_1N_2$ in terms of $N_1$ smaller DFTs of sizes $N_2$ to reduce the computation time to \O{N \log N}. 

An alternative approach is to simply apply the DFT matrix to the input sequence. The DFT matrix, $W$, is a Vandermonde matrix for the roots of unity up to a normalization factor:
\begin{equation}
    \label{eq:vandermonde}
    W_{nk} = \left({e^{-{\frac{2\pi i}{N}} n k}}/{\sqrt{N}}\right),
\end{equation}
where  $n,k = 0, \ldots , N-1$.
This matrix multiplication is an \O{N^2} operation, which has higher asymptotic complexity than the FFT, but turns out to be faster for relatively shorter sequences on TPUs.

\subsection{FNet architecture}
\label{subsec:architecture}

\begin{figure}
    \centering
    \includegraphics[width=0.38\textwidth]{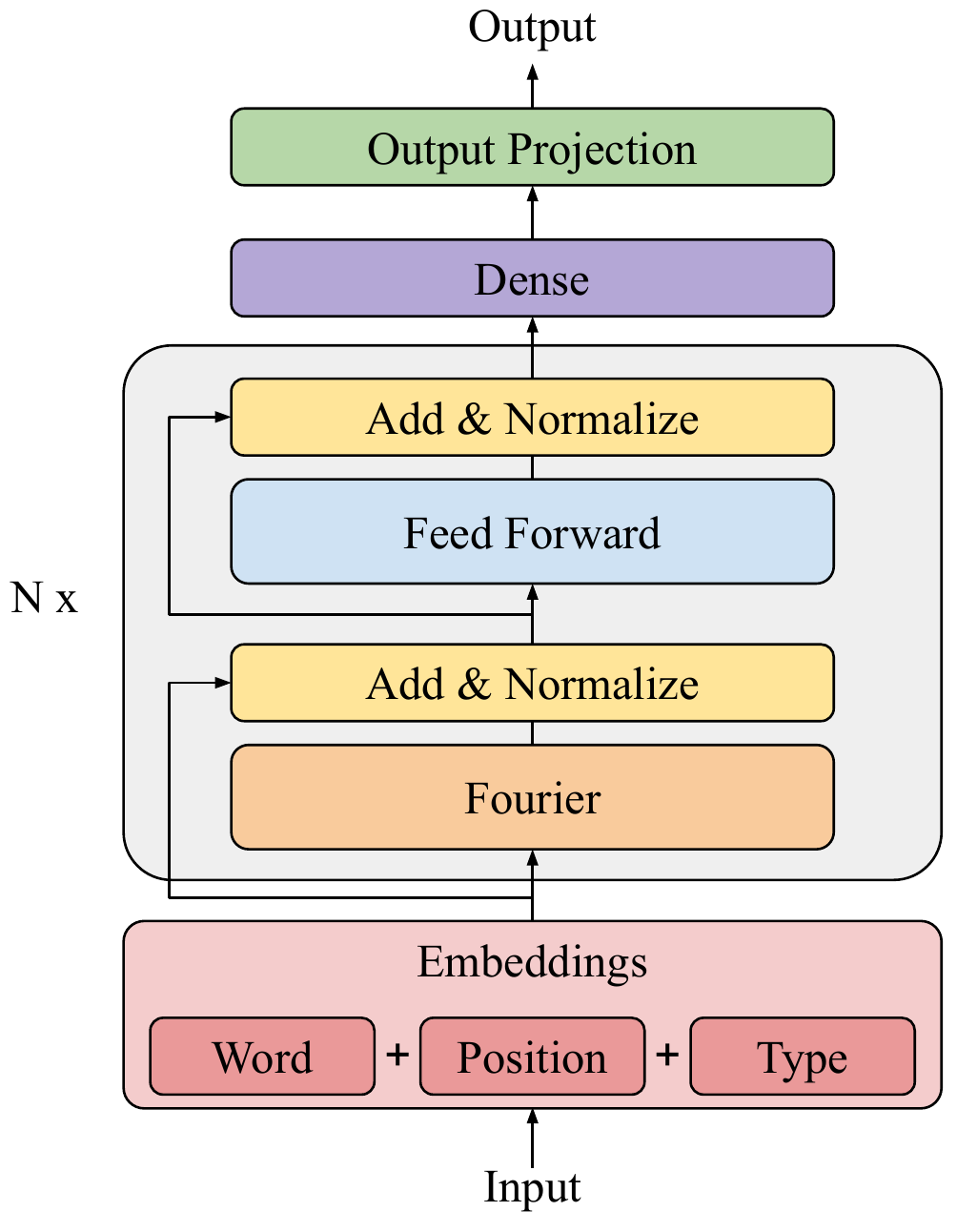}
    \caption{FNet architecture with $N$ encoder blocks.}
    \label{fig:architecture}
\end{figure}

FNet is an attention-free Transformer architecture, wherein each layer consists of a Fourier mixing sublayer followed by a feed-forward sublayer. The architecture is shown in Figure \ref{fig:architecture}. Essentially, we replace the self-attention sublayer of each Transformer encoder layer with a Fourier sublayer, which applies a 2D DFT to its (sequence length, hidden dimension) embedding input -- one 1D DFT along the sequence dimension, $\mathcal{F}_{\textrm{seq}}$, and one 1D DFT along the hidden dimension, $\mathcal{F}_{\textrm{h}}$:\footnote{The relative ordering of $\mathcal{F}_{\textrm{seq}}$ and $\mathcal{F}_{\textrm{h}}$ in Equation \eqref{eq:fourier_layer} is immaterial because the two 1D DFTs commute.}
\begin{equation}
    \label{eq:fourier_layer}
    y = \Re\left(\mathcal{F}_{\textrm{seq}}\left(\mathcal{F}_{\textrm{h}} (x)\right) \right) .
\end{equation}
As indicated by Equation \eqref{eq:fourier_layer}, we only keep the real part of the result; hence, we do not need to modify the (nonlinear) feed-forward sublayers or output layers to handle complex numbers. We found that FNet obtained the best results when the real part of the total transformation was only extracted at the end of the Fourier sublayer; that is, after applying both $\mathcal{F}_{\textrm{seq}}$ and $\mathcal{F}_{\textrm{h}}$.
We also experimented with the Hadamard, Hartley and Discrete Cosine Transforms. Of these three, the Hartley Transform was the strongest alternative, obtaining comparable accuracy to Equation \eqref{eq:fourier_layer}; see Appendix \ref{app:things_we_tried} for details.

The simplest interpretation for the Fourier Transform is as a particularly effective mechanism for mixing tokens, which  provides the feed-forward sublayers sufficient access to all tokens. Because of the duality of the Fourier Transform, we can also view each alternating encoder block as applying alternating Fourier and inverse Fourier Transforms, transforming the input back and forth between the ``time'' and frequency domain. Because multiplying by the feed-forward sublayer coefficients in the frequency domain is equivalent to convolving (with a related set of coefficients) in the time domain, FNet can be thought of as alternating between multiplications and  
convolutions.\footnote{This is merely an intuition; the reality is more complicated due to the presence of residual connections and since the transformation in Equation \eqref{eq:fourier_layer} is no longer invertible if we only use the real component.}

We use the same embedding layers as in \citet{devlin2018bert}; namely, we combine the word embeddings, absolute position embeddings of the tokens and type embeddings of the sentences. Because of the positional information encoded by the Fourier Transform in Equation \eqref{eq:dft} (see $n$, $k$ indices), FNet performs just as well without position embeddings. Nevertheless, we include the position embeddings to allow for a cleaner comparison with BERT.

\subsection{Implementation}
\label{subsec:implementation}

Empirically, we found that on GPUs: the FFT is faster than matrix multiplications for all sequence lengths we consider ($512 - 8192$ tokens), whereas on TPUs: for relatively shorter sequences ($\leq 4096$ tokens), it is faster to cache the DFT matrix and then compute the DFT through matrix multiplications than using the FFT; for longer sequences, the FFT is faster. As a result, our GPU FNet implementation always uses the FFT, while our TPU implementation computes the 2D DFT using matrix multiplications for sequences up to lengths of $4096$ and the FFT for longer lengths.
Presumably the GPU vs TPU difference is primarily a result of two factors: (1) TPUs are even more highly optimized for matrix multiplications than GPUs, and (2) GPUs offer a more efficient FFT implementation than TPUs. We suspect that FNet will only become more performant on TPUs as the TPU implementation of the FFT improves. Our model uses JAX and, in particular, the Flax framework\footnote{\url{https://github.com/google/flax}}. Core model code is given in Appendix \ref{app:code} and the full source core is available online.\footnote{\url{https://github.com/google-research/google-research/tree/master/f_net}}
\section{Results}
\label{sec:results}

\subsection{Transfer learning}
\label{subsec:transfer_learning}

\begin{table}
    \caption{Number of mixing layer operations (forward pass) and learnable parameters, excluding any task specific output projection layers. $n$ is the sequence length and $d_{h}$ is the model hidden dimension. The mixing layer operations are given on a per layer basis.}
    \label{tab:model_params}
    \centering
    \setlength{\tabcolsep}{3pt}
    \begin{tabular}{l | c | c  c}
        \hline
         & Mixing layer ops & \multicolumn{2}{c}{Model params} \\ 
         Model & (per layer) & Base & Large \\  \hline \hline
         BERT & $2 n^2 d_{h} + 4 n d_{h}^2$ & $112$M & $339$M \\
         Linear & $n^2 d_{h} + n d_{h}^2$ & $94$M & $269$M \\
         FNet (mat) & $n^2 d_{h} + n d_{h}^2$ & $83$M & $238$M \\
         FNet (FFT) & $n d_{h} \log(n) +$ & $83$M & $238$M \\
         & $n d_{h}\log(d_{h})$ & \\
         Random & $n^2 d_{h} + nd_{h}^2$ & $83$M & $238$M \\
         FF-only & $0$ & $83$M & $238$M \\ \hline
    \end{tabular} 
\end{table}

\begin{table*}[tb]
    \caption{GLUE Validation results on TPUs, after finetuning on respective tasks. We report the mean of accuracy and F1 scores for QQP and MRPC, Spearman correlations for STS-B and accuracy scores for all other tasks. The MNLI metrics are reported by the match/mismatch splits.
    Average scores exclude any failure cases. After controlling for batch size and training steps, the GPU metrics (not shown) are similar.}
    \label{tab:glue}
    \centering
    \begin{tabular}{l| c c c c c c c c | c}
        \hline
         Model  & MNLI & QQP & QNLI & SST-2 & CoLA & STS-B & MRPC & RTE & Avg. \\ \hline \hline
         BERT-Base & \textbf{84/81} & \textbf{87} & \textbf{91} & 93 & 73 & \textbf{89} & \textbf{83} & \textbf{69} & \textbf{83.3} \\
         Linear-Base & 74/75 & 84 & 80 & 94 & 67 & 67 & 83 & 69 & 77.0 \\
         FNet-Base & 72/73 & 83 & 80 & \textbf{95} & 69 & 79 & 76 & 63 & 76.7 \\
         Random-Base & 51/50 & 70 & 61 & 76 & 67 & 4 & 73 & 57 & 56.6 \\
         FF-only-Base & 34/35 & 31 & 52 & 48 & 67 & FAIL & 73 & 54 & \ 49.3 \\ 
         FNet-Hybrid-Base & 78/79 & 85 & 88 & 94 & \textbf{76} & 86 & 79 & 60 & 80.6 \\ \hline
         BERT-Large & \textbf{88/88} & \textbf{88} & \textbf{92} & \textbf{95} & 71 & \textbf{88} & 86 & 66 & \textbf{84.7} \\
         Linear-Large & 35/36 & 84 & 80 & 79 & 67 & 24 & 73 & 60 & 59.8 \\
         FNet-Large & 78/76 & 85 & 85 & 94 & 78 & 84 & \textbf{88} & 69 & 81.9 \\
         FNet-Hybrid-Large & 79/80 & 87 & 89 & 92 & \textbf{81} & \textbf{88} & 86 & \textbf{70} & 83.6 \\ \hline
    \end{tabular}
\end{table*}

We compare FNet and Transformer architectures in a common transfer learning setting. For a fuller picture, we compare multiple models (see Table \ref{tab:model_params} for parameter counts in ``Base'' configuration): 
\begin{itemize}
    \item BERT-Base: a Transformer encoder model.
    \item FNet encoder: we replace every self-attention sublayer with a Fourier sublayer.
    \item Linear encoder: we replace each self-attention sublayer with a two learnable, dense, linear sublayers, one applied to the hidden dimension and one to the sequence dimension.
    \item Random encoder: we replace each self-attention sublayer with a two constant random matrices, one applied to the hidden dimension and one applied to the sequence dimension.
    \item Feed Forward-only (FF-only) encoder: we completely remove the self-attention sublayer; so that this model has no token mixing.
\end{itemize}

Despite its simplicity, the Linear baseline turns out to be surprisingly accurate and fast. Our Linear model is similar to the MLP-Mixer \citep{tolstikhin2021mlp} (for vision) and also the Random Synthesizer \citep{tay2020synthesizer}, but simplifies the latter model further by removing the multiple heads and softmax projections, resulting in just two matrix multiplications in the mixing sublayer. 

\begin{figure*}[tb]
    \centering
    \includegraphics[width=\textwidth]{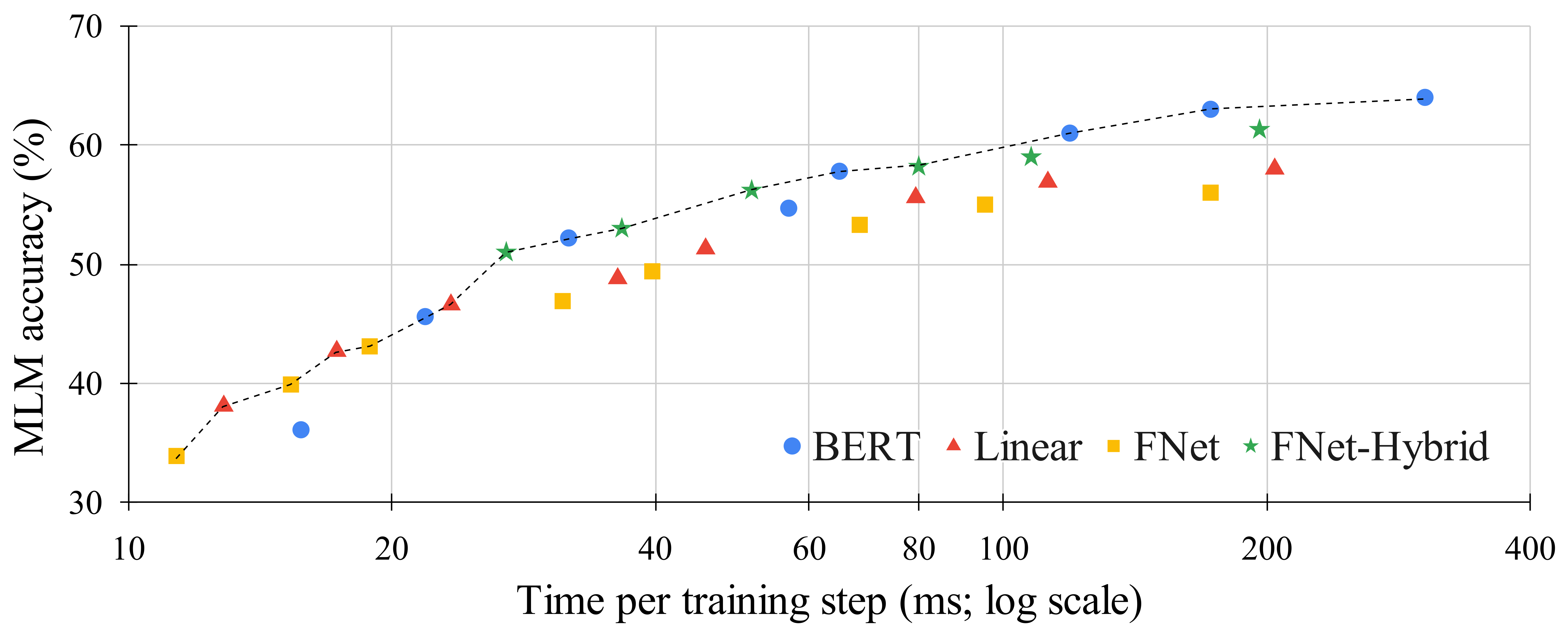}
    \caption{
    Speed-accuracy trade-offs for GPU pre-training. The dashed line shows the Pareto efficiency frontier, indicating the best trade-offs. For smaller models (faster training speeds; left-hand side of figure), the FNet (yellow squares) and Linear (red triangles) models define the frontier, while for larger models (slower training speeds; right-hand side of figure), BERT (blue circles) and FNet-Hybrid (green stars) define the frontier.}
    \label{fig:gpu_speed_accuracy}
\end{figure*}

It is reasonable to expect that the Linear encoder, which uses densely parameterized mixing layers, will learn more flexibly than FNet, which uses parameter-free mixing layers. As we will show, although the Linear-Base model outperforms FNet-Base slightly on GLUE ($0.3$ points), it has several efficiency drawbacks relative to FNet: it has a much larger memory footprint (see Table \ref{tab:lra_training_speed}), it is slower to train on regular $512$ sequence lengths (see Table \ref{tab:pretrain_speed}), and scales significantly worse on long sequence lengths (see Tables \ref{tab:lra_training_speed}-\ref{tab:lra_inference_speeds}).\footnote{On the other hand, the smaller sized Linear models do generally perform well on $512$ sequence lengths; see Figure \ref{fig:gpu_speed_accuracy}.} We also found that Linear-Large was more difficult to train due to gradient blow up (see ``Large'' scores in Table \ref{tab:glue}).

\begin{table*}[tb]
    \caption{Pre-training and inference speeds in milliseconds per batch of 64 examples on GPU ($8$ V100 chips) and 256 examples on TPU ($4\times4$ v3 chips), alongside GFLOPS for a forward pass of a single example. Speed-up multipliers relative to BERT are given in parentheses.}
    \label{tab:pretrain_speed}
    \centering
    \begin{tabular}{l | c c | c c | c}
        \hline
         & \multicolumn{2}{c|}{Pre-training} & \multicolumn{2}{c|}{Inference} & GFLOPS  \\  
         Model & GPU & TPU & GPU & TPU & /example \\ \hline \hline
         BERT-Base & 305 & 213 & 82  & 32 & 98 \\
         Linear-Base & 199 (1.5x) & 149 (1.4x) & 52 (1.6x)  & 20 (1.6x) & 71 (73\%) \\
         FNet-Base & 169 (1.8x) & 128 (1.7x) & 46 (1.8x) & 23 (1.4x) & 62 (63\%) \\
         Random-Base & 182 (1.7x) & 130 (1.6x) & 52 (1.6x) & 22 (1.4x) & 71 (73\%) \\
         FF-only-Base & \textbf{162 (1.9x)} & \textbf{118 (1.8x)} & \textbf{43 (1.9x)} & \textbf{16 (2.0x)} & \textbf{59 (60\%)} \\
         FNet-Hybrid-Base & 198 (1.5x) & 149 (1.4x) & 51 (1.6x) & 24 (1.3x) & 68 (69\%) \\ \hline
         BERT-Large & OOM & 503 & 263 & 111 & 337 \\
         Linear-Large & 592 & 397 (1.3x) & 170 (1.5x) & 108 (1.0x) & 247 (73\%) \\
         FNet-Large & \textbf{511} & \textbf{275 (1.8x)} & \textbf{149 (1.8x)} & \textbf{82 (1.4x)} & \textbf{217 (64\%)} \\ 
         FNet-Hybrid-Large & 541 & 294 (1.7x) & 157 (1.7x) & 84 (1.3x) & 227 (67\%)\\ \hline
    \end{tabular}
\end{table*}

We adopt the same fixed ``Base'' and ``Large'' model and training configurations as for the original BERT \citep{devlin2018bert}, except that we pre-train on the much larger C4 dataset \citep{raffel2019exploring} and use a $32000$ SentencePiece vocabulary model \citep{kudo2018sentencepiece} (see Appendix \ref{app:mlm} for full pre-training details). For fine-tuning on the GLUE benchmark \citep{wang2018glue}, we found that different BERT runs with the same base learning rate could yield slightly different results. Consequently, for the Base (Large) models, we performed 3 (6) trials, respectively, for each base learning rate and reported the best result across all experiments. This reflects our observation that BERT-Large was less stable than BERT-Base, as noted in \citet{devlin2018bert}.

We report the results for the best base learning rate (no early stopping) on the GLUE Validation split in Table \ref{tab:glue}.\footnote{WNLI is excluded in \citet{devlin2018bert}. BERT’s accuracy on WNLI is below baseline, unless a special training recipe is used. See also (12) in \url{https://gluebenchmark.com/faq.}}
For Base models, results mirror the pre-training metrics (see Appendix \ref{app:mlm}): BERT performs best. FNet and the Linear model both  underperform BERT by $7.5-8\%$. Referring to Table \ref{tab:pretrain_speed}, we see that although less accurate, FNet trains significantly faster than BERT -- $80\%$ faster on GPUs and $70\%$ faster on TPUs -- and performs $63\%$ of BERT's FLOPS. Measured in isolation, the Fourier sublayers perform forward and backward passes an order of magnitude faster than the self-attention sublayers (see Appendix \ref{app:mixing_speeds}), but FNet's overall training speed is impeded by the feed-forward sublayers that all models share.

Returning to Table \ref{tab:glue}: the FF-only model severely underperforms all other models: as expected, token mixing is critical to the expressivity of the model. For example, $50\%$ accuracy scores on the binary classification tasks (QNLI, SST-2, RTE), indicate that the model fails to learn the tasks. The weak accuracy of the Random model suggests that not just any mixing will do; rather, a structured mixing is required. We also include metrics from a hybrid FNet attention model. In the hybrid model, we replace the final two Fourier sublayers of FNet with self-attention sublayers -- other configurations are possible, but we generally found that replacing the final layers worked best; see Appendix \ref{app:hybrid_ablations}. With the addition of just two self-attention sublayers, the hybrid FNet models achieve $97\%$ and $99\%$ of their respective BERT counterpart's accuracies with only limited speed degradations (see Table \ref{tab:pretrain_speed}).

Interestingly, the gap between BERT and FNet shrinks to just $3\%$ for Large models; this is likely due to FNet-Large being more stable during training than BERT-Large.\footnote{\citet{devlin2018bert} obtain a roughly $2.5$ average point boost on the \emph{Test} split going from BERT-Base to BERT-Large. We only see a roughly $1.5$ boost on the \emph{Validation} split, which may be due to reduced headroom.} The Linear-Large model severely underperforms its Base counterpart on GLUE benchmark due to training instabilities. We generally found that the Linear model and BERT were less stable than the models with no parameters in their mixing sublayers, namely the FNet, Random and FF-only models.

The speed vs MLM accuracy curve for GPU ($8$ V100 chips) pre-training is shown in Figure \ref{fig:gpu_speed_accuracy} (see Appendix \ref{app:tpu} for TPU results). Both TPU and GPU models are trained for $1$ million steps as in \citet{devlin2018bert}. Motivated by the models considered in \citet{turc2019well}, we evaluated several model sizes; see Table \ref{tab:model_sizes} in Appendix \ref{app:mlm}. We found that the smaller model architectures benefited from larger learning rates, so we select the best result using $10^{-3}$ and $10^{-4}$ for all models.\footnote{We have opted to compare FNet with Transformer models as the latter are the most commonly used models in NLP transfer learning settings. It would also be interesting to compare FNet with convolutional-based models, although, to our knowledge, such models have only recently found limited success in pre-training NLP setups \citep{tay2021pre}; and even there, the authors did not consider the small model regime.}

\begin{table*}[tb]
    \caption{Accuracy, inference speed and memory usage results on the Long-Range Arena (LRA) benchmark.}
    \label{tab:lra}
    \begin{subtable}[t]{\textwidth}
    \centering
    \begin{tabular}{l | c c c c c c | c}
        \hline
         Model & ListOps & Text & Retrieval & Image & Pathfinder & Path-X & Avg.  \\ \hline \hline
         Transformer (ours) & {36.06} & 61.54 & \textbf{59.67} & {41.51} & 80.38 & OOM & \textbf{55.83}  \\ 
         Linear (ours) & 33.75 & 53.35 & 58.95 & 41.04 & \textbf{83.69} & FAIL & 54.16 \\
         FNet (ours) & 35.33 & {65.11} & 59.61 & 38.67 & 77.80 & FAIL & 55.30  \\ \hline
         Transformer (*) & 36.37 & 64.27 & 57.46 & 42.44 & 71.40 & FAIL & \underline{54.39} \\
         Local Attention (*) & 15.82 & 52.98 & 53.39 & 41.46 & 66.63 & FAIL & 46.06 \\
         Sparse Trans. (*) & 17.07 & 63.58 & \textbf{59.59} & \textbf{44.24} & 71.71 & FAIL & 51.24 \\
         Longformer (*) & 35.63 & 62.85 & 56.89 & 42.22 & 69.71 & FAIL & 53.46 \\
         Linformer (*) & 35.70 & 53.94 & 52.27 & 38.56 & \underline{76.34} & FAIL & 51.36 \\
         Reformer (*) & \textbf{37.27} & 56.10  & 53.40 & 38.07 & 68.50 & FAIL & 50.67 \\
         Sinkhorn Trans. (*) & 33.67 & 61.20 & 53.83 & 41.23 & 67.45 & FAIL & 51.39 \\
         Synthesizer (*) & \underline{36.99} & 61.68 & 64.67 & 41.61 & 69.45 & FAIL & 52.88 \\
         BigBird (*) & 36.05 & 64.02 & \underline{59.29} & 40.83 & 74.87 & FAIL & \textbf{55.01} \\
         Linear Trans. (*) & 16.13 & \textbf{65.90} & 53.09 & 42.34 & 75.30 & FAIL & 50.55 \\
         Performer (*) & 18.01 & \underline{65.40} & 53.82 & \underline{42.77} & 77.05 & FAIL & 51.41  \\ \hline
    \end{tabular}
    \caption{Accuracy results obtained on TPUs as in \citet{tay2020long}. Asterisked results quoted from \citet{tay2020long}. Average does not include the Path-X task, which all models fail (Transformer due to memory limits; others perform no better than chance).}
    \label{tab:lra_accuracy}
    \end{subtable}
    \begin{subtable}[t]{\textwidth}
    \centering
    \setlength{\tabcolsep}{4pt}
    \begin{tabular}{l | c c c c c | c c c c c}
        \hline
         & \multicolumn{5}{c|}{Training Speed (steps/s)} & \multicolumn{5}{c}{Peak Memory Usage (GB)} \\ 
         Seq. length & 512 & 1024 & 2048 & 4096 & 8192 & 512 & 1024 & 2048 & 4096 & 8192  \\ \hline \hline
         Transformer & 21 & 10 & 4 & OOM & OOM  & 1.6 & 4.0 & 12.2 & OOM & OOM \\
         Linear & 34 (1.6x) & 19 (1.8x) & 9 (2.0x) & 4 & OOM & 0.9 & 1.6 & 2.8 & 6.9 & OOM  \\
         FNet (FFT) &\textbf{43 (2.0x)} & \textbf{24 (2.3x)} & \textbf{14 (3.2x)} & \textbf{7} & \textbf{4} & \textbf{0.8} & \textbf{1.3} & \textbf{2.2} & \textbf{3.9} & \textbf{7.4}  \\
         Performer & 28 (1.3x) & 15 (1.5x) & 9 (1.9x) & 4 & 2 & 1.1 & 1.9 & 3.1 & 5.5 & 10.4 \\ \hline
    \end{tabular}
    \caption{GPU training for sequence lengths up to 8192. Only the fastest efficient Transformer, namely Performer, from \citet{tay2020long} is shown. Left: training speeds (in steps per second; larger is better), with speed-up multipliers relative to the Transformer given in parentheses. Right: peak memory usage (in GB; smaller is better).}
    \label{tab:lra_training_speed}
    \end{subtable}
    \begin{subtable}[t]{\textwidth}
    \centering
    \begin{tabular}{l | c c c c c c}
        \hline
         Seq. length & 512 & 1024 & 2048 & 4096 & 8192 & 16384 \\ \hline \hline
         Transformer & 12 & 28 & 76 & 244 & OOM & OOM \\
         Linear & 9 (1.4x) & 14 (2.0x) & 30 (2.6x) & 72 (3.4x) & 208 & OOM \\
         FNet (FFT) & \textbf{8 (1.5x)} & \textbf{12 (2.3x)} & \textbf{23 (3.4x)} & \textbf{43 (5.7x)} & \textbf{83} & \textbf{164} \\
         Performer & 11 (1.2x) & 17 (1.6x) & 32 (2.4x) & 60 (4.0x) & 116 & 238 \\ \hline
    \end{tabular}
    \caption{GPU inference speeds on the LRA Text classification task (in milliseconds per batch; smaller is better). Only the fastest efficient Transformer, Performer, from \citet{tay2020long} is shown. Speed up relative to the Transformer is given in parentheses.}
    \label{tab:lra_inference_speeds}
    \end{subtable}
\end{table*}

The GPU (Figure \ref{fig:gpu_speed_accuracy}), and TPU (Figure \ref{fig:tpu_speed_accuracy} in Appendix \ref{app:tpu}) results display the same trends. For larger, slower models, BERT and FNet-Hybrid define the Pareto speed-accuracy efficiency frontier. For smaller, faster models, FNet and the Linear model define the efficiency frontier.

\subsection{Long-Range Arena (LRA) benchmark}
\label{subsec:lra}

Of the efficient Transformers evaluated on LRA benchmark by \citet{tay2020long}, their results suggest that (1) the vanilla Transformer is (by a small margin) the second most accurate model, and (2) the Performer \citep{choromanski2020rethinking} is the fastest model.  We benchmark FNet's accuracy against both of these models using \citet{tay2020long}'s codebase and running on the same hardware ($4\times4$ TPU v3 chips); the results are shown in Table \ref{tab:lra_accuracy}.\footnote{The ``Linear'' model in Table \ref{tab:lra} is the baseline model introduced in Section \ref{subsec:transfer_learning}.} To ensure a fair comparison, we also report the results of our own experiments for the vanilla Transformer (see Appendix \ref{app:lra} for details). 

Table \ref{tab:lra_accuracy} suggests that, in aggregate, the (vanilla) Transformer and FNet obtain comparable results. Given that the Transformer is the second most accurate model evaluated by \citet{tay2020long} and that the relative differences in the average accuracy scores within Table \ref{tab:lra_accuracy} are small, our results suggest that FNet is competitive with the most accurate of the efficient Transformers on LRA.

Turning to efficiency, in Table \ref{tab:lra_training_speed}, we provide training speed and memory usage statistics from our experiments on GPUs ($8$ V100 chips); see Appendix \ref{app:tpu} for results on TPUs. We perform a sweep over sequence lengths $\{512, 1024, 2048, 4096, 8192\}$. On GPUs, FNet is much faster than all other models across all sequence lengths, due to the highly efficient FFT implementation on GPUs.  Table \ref{tab:lra_training_speed} also indicates that FNet has a lighter memory footprint (this holds for both GPUs and TPUs; see extended results in Appendix \ref{app:tpu}). This is partly because FNet has no learnable parameters in its mixing sublayer, but also due to the FFT's efficiency, especially at longer sequence lengths. Lastly, Table \ref{tab:lra_inference_speeds} shows that training speed gains generally carry over to inference gains (see Appendix \ref{app:tpu} for detailed TPU results).

\section{Conclusions}
\label{sec:conclusion}

In this work, we studied simplified token mixing modules for Transformer-like encoder architectures, making several contributions. First, we showed that simple, linear mixing transformations, along with the nonlinearities in feed-forward layers, can competently model diverse semantic relationships in text. Second, we introduced FNet, a Transformer-like model wherein the self-attention sublayer is replaced by an unparameterized Fourier Transform. FNets achieve $92$ and $97\%$ of their respective BERT-Base and BERT-Large counterparts' accuracy on the GLUE benchmark, but train $70-80\%$ faster on GPUs/TPUs. Third, because of its favorable scaling properties, FNet is very competitive with the ``efficient Transformers'' evaluated on the Long-Range Arena benchmark, matching the accuracy of the most accurate models while being much faster and lighter on memory. 

Our work highlights the potential of linear units as a drop-in replacement for the attention mechanism in text classification tasks. We found the Fourier Transform to be a particularly efficient and effective mixing mechanism, due to the speed of the FFT. However, we only performed a cursory survey of other linear transformations (see also Appendix \ref{app:things_we_tried}), and additional fast alternatives are worth exploring.

Given the speed and accuracy advantages of smaller FNet models relative to Transformers, we suspect that FNet will be effective as a lightweight, distilled student model deployed in resource-constrained settings such as production services or on edge devices. The need for such lightweight serving models is only forecast to grow given the interest in giant models \citep{raffel2019exploring, brown2020language, lepikhin2020gshard}. A natural avenue to explore in this regard is knowledge distillation of small FNet models from larger Transformer teacher models, following, for example, \citet{sanh2019distilbert, jiao2019tinybert, turc2019well}.

Another aspect of interest and worthy of further study is hybrid FNet-attention models. We found that adding only a few self-attention sublayers to FNet offers a simple way to trade speed for accuracy. Specifically, replacing the final two Fourier sublayers with self-attention provided $97-99\%$ of BERT's accuracy with limited speed penalties.

Throughout this work we have restricted our focus to encoders. FNet decoders can be designed by ``causally'' masking the Vandermonde matrix, but a lower level implementation is required to introduce causal masking to FFTs. How to adapt Fourier mixing for encoder-decoder cross-attention is an open question as evidence suggests that cross-attention may be crucial to performance \citep{you2020hard}. We have focused on tasks which do not require generation so we leave FNet decoders and encoder-decoder setups to future work; although we do remark that the FNet encoder could be used as a drop in replacement in a Transformer as other works have successfully demonstrated; see, for example, \cite{zaheer2020big, guo2021longt5}.


\bibliographystyle{acl_natbib}
\bibliography{anthology, references}

\appendix
\clearpage

\section{Appendices}
\label{sec:appendix}

\subsection{Pre-training details}
\label{app:mlm}

We adopt the same fixed ``Base'' and ``Large'' model and learning configurations as for the original BERT \citep{devlin2018bert}. We train on the much larger C4 dataset \citep{raffel2019exploring} and use a $32000$ SentencePiece  vocabulary model \citep{kudo2018sentencepiece} trained on a $100$ million sentence subset of C4. Our TPU experiments use a batch size of $256$ as in \citet{devlin2018bert} and are each run on $4\times4$ TPU v3 chips. Our GPU experiments use a smaller batch size of $64$ and are run on $8$ V100 chips. Because the training configuration is lifted from \citet{devlin2018bert}, it may be slightly biased towards the BERT attention model.

\begin{table}
    \caption{Loss and accuracy pre-training metrics on TPUs. The GPU metrics are very similar. ``B'' denotes Base, ``L'' is Large and ``H'' is Hybrid.}
    \label{tab:pretraining_metrics}
    \setlength{\tabcolsep}{4.5pt}
    \centering
    \begin{tabular}{l| c c c | c c}
        \hline
         & \multicolumn{3}{c|}{Loss} & \multicolumn{2}{c}{Accuracy} \\ 
         Model & Total & MLM & NSP & MLM & NSP \\ \hline \hline
         BERT-B & \textbf{1.76} & \textbf{1.48} & \textbf{0.28} & \textbf{0.68} & \textbf{0.86} \\ 
         Linear-B & 2.12 & 1.78 & 0.35 & 0.62 & 0.83 \\ 
         FNet-B & 2.45 & 2.06 & 0.40 & 0.58 & 0.80 \\
         Random-B & 5.02 & 4.48 & 0.55 & 0.26 & 0.70 \\ 
         FF-only-B & 7.54 & 6.85 & 0.69 & 0.13 & 0.50 \\
         FNet-H-B & 2.13 & 1.79 & 0.34 & 0.63 & 0.84 \\ \hline
         BERT-L & \textbf{1.49} & \textbf{1.23} & \textbf{0.25} & \textbf{0.72} & \textbf{0.88} \\ 
         Linear-L & 1.91 & 1.60 & 0.31 & 0.65 & 0.85 \\ 
         FNet-L & 2.11 & 1.75 & 0.36 & 0.63 & 0.82 \\
         FNet-H-L & 1.89 & 1.58 & 0.31 & 0.67 & 0.85 \\\hline
    \end{tabular}
\end{table}

Table \ref{tab:pretraining_metrics} summarizes the pre-training metrics for the different models; the pre-training speeds are shown in Table \ref{tab:pretrain_speed} in the main text. Although they have weaker accuracy metrics, the Linear model and FNet train nearly $80\%$ faster than BERT on GPUs, and $70\%$ faster on TPUs (see Table \ref{tab:pretrain_speed}).  We also find that the three models with no learnable parameters in their mixing layer, namely FNet, the  Random model and the FF-only model, are the most stable during training.

BERT's higher accuracy on the MLM pre-training task is not simply a result of having more parameters than the other models. Indeed, Table \ref{tab:pretraining_metrics} shows that BERT-Base is actually more accurate than FNet-Large, which contains more than twice as many parameters. BERT is presumably more expressive because the mixing (attention) weights are both task specific and token dependent, determined by token-token (query-key) dot products; see also \citet{tay2020synthesizer}. FNet's mixing weights, on the other hand, are neither task specific nor token dependent.

\begin{table}
    \caption{Pre-training model sizes (ignoring output projection layers). As in \citet{turc2019well}, for all models, we fix the feed-forward size to $4d_{h}$ and the number of self-attention heads to $d_{h}/64$. Smaller architectures have a similar number of parameters across all models because the majority of parameters are in the embedding layers. Each FNet-Hybrid (``FNet-H'') model contains 2 self-attention sublayers. We exclude FNet-Hybrid models with only 2 total layers.}
    \label{tab:model_sizes}
    \centering
    \setlength{\tabcolsep}{4pt}
    \begin{tabular}{c c | c c c c}
        \hline
        \multicolumn{2}{c|}{Dimensions} & \multicolumn{4}{c}{Parameters (millions)}  \\ 
        $d_{h}$ & Layers & BERT & Linear & FNet & FNet-H  \\
          \hline \hline
        768  & 12 & 111 & 93 & 83 & 88 \\
        512  & 12 & 55 & 49 & 42 & 44 \\
        512  & 8 & 42 & 38 & 34 & 36 \\
        256 & 8 & 15 & 15 & 13 & 13 \\
        512 & 4 & 30 & 28 & 26 & 28 \\
        256  & 4 & 12 & 12 & 11 & 11\\
        256 & 2 & 10 & 10 & 10 & - \\
        128 & 2 & 5 & 5 & 4 & - \\ \hline
    \end{tabular}
\end{table}

\begin{figure*}
    \centering
    \includegraphics[width=\textwidth]{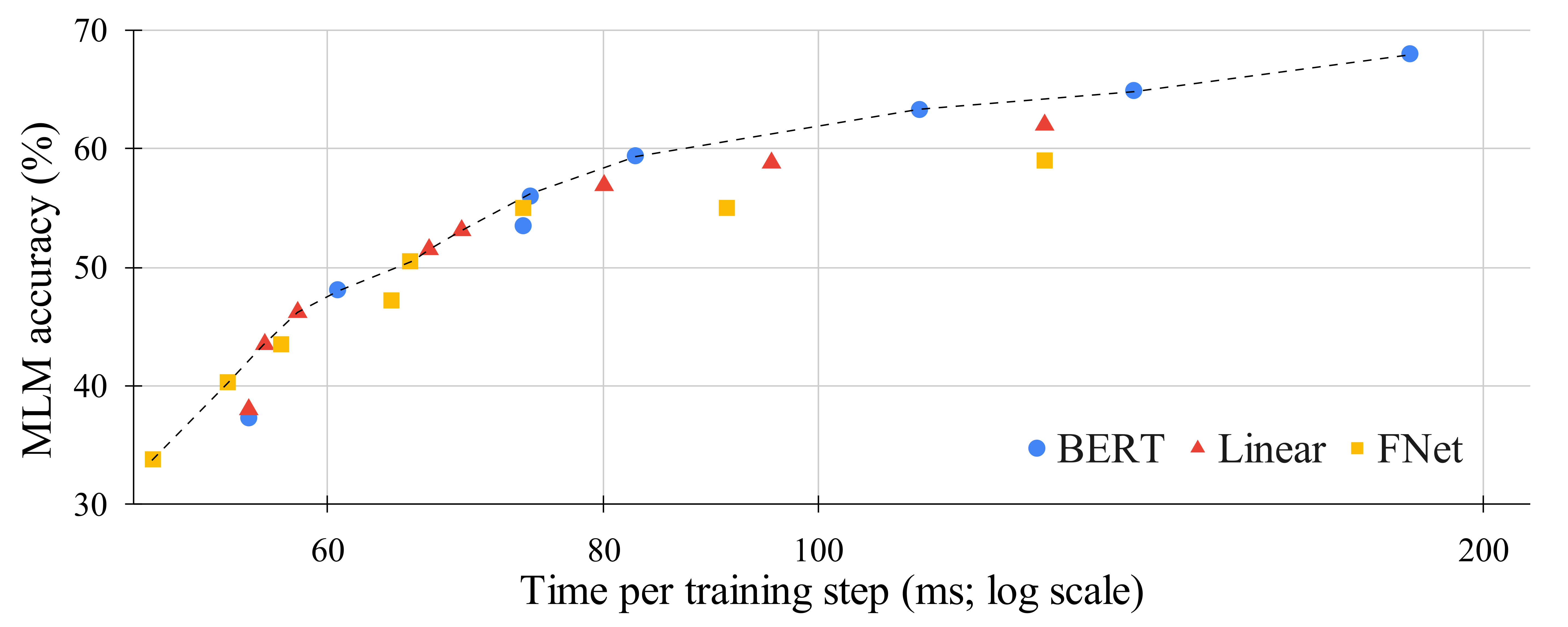}
    \caption{Speed-accuracy trade-offs for TPU pre-training. The dashed line shows the Pareto efficiency frontier.}
    \label{fig:tpu_speed_accuracy}
\end{figure*}

Finally, Table \ref{tab:model_sizes} shows the model sizes that were used to construct Figure \ref{fig:gpu_speed_accuracy} (main text) and Figure \ref{fig:tpu_speed_accuracy} (Appendix \ref{app:tpu}).

\begin{table*}
    \caption{TPU training speeds (in steps per second; larger is better), inference speeds (in milliseconds per batch; smaller is better) and peak memory usage during training (in GB; smaller is better) on the Long-Range Arena Text classification task. Speed up multipliers relative to the Transformer are given in parentheses.}
    \label{tab:lra_training_speed_tpu}
    \centering
    \begin{tabular}{l | c c c c c c}
        \hline
         Seq. length & 512 & 1024 & 2048 & 4096 & 8192 & 16386  \\ \hline \hline
         & \multicolumn{6}{c}{Training Speed (steps/s)} \\ \hline
         Transformer & 8.0 & 5.6 & 1.7 & OOM & OOM & OOM \\
         Linear & 9.4 (1.2x) & \textbf{9.1 (1.6x)} & \textbf{7.6 (4.5x)} & 3.9 & 1.4 & OOM \\
         FNet (mat) & \textbf{9.5 (1.2x)} & \textbf{9.1 (1.6x)} & 6.1 (3.6x) & 3.0 & 0.8 & 0.2 \\
         FNet (FFT) & 8.6 (1.1x) & 6.0 (1.1x) & 3.2 (1.9x) & 1.6  & 0.8 & 0.3 \\
         Performer & 9.2 (1.2x) & 8.4 (1.5x) & 6.9 (4.1x) & \textbf{4.2} & \textbf{2.2} & \textbf{1.1} \\ \hline
         & \multicolumn{6}{c}{Inference Speed (ms/batch)} \\ \hline
         Transformer & 7.0 & 13.2 & 39.4 & 129.9 & 490.2 & OOM \\
         Linear & \textbf{5.6 (1.2x)} & \textbf{6.5 (2.0x)} & \textbf{9.6 (4.1x)} & 20.4 (6.4x) & 54.6 (9.0x) & OOM \\
         FNet (mat) & 6.0 (1.2x) & 7.7 (1.7x) & 15.4 (2.6x) & 40.7 (3.2x) & 137.0 (3.6x) & 454.5 \\
         FNet (FFT) & 10.8 (0.7x) & 16.8 (0.8x) & 29.9 (1.3x) & 58.8 (2.2x) & 113.6 (4.3x) & 263.2 \\
         Performer & 6.1 (1.2x) & 7.2 (1.8x) & 10.1 (3.9x) & \textbf{17.5 (7.4x)} & \textbf{31.8 (15.4x)} & \textbf{61.0} \\ \hline
         & \multicolumn{6}{c}{Peak Memory Usage (GB)} \\ \hline
         Transformer & 1.1 & 2.1 & 5.8 & 9.1 & OOM & OOM \\
         Linear & 0.9 & 1.1 & 1.9 & 4.9 & 14.8 & OOM \\
         FNet (mat) & \textbf{0.8} & \textbf{0.9} & \textbf{1.3} & 2.2 & 4.8 & 11.9 \\
         FNet (FFT) & \textbf{0.8} & \textbf{0.9} & \textbf{1.3} & \textbf{2.0} & \textbf{3.5} & \textbf{6.3} \\
         Performer & 1.0 & 1.3 & 1.8 & 3.0 & 5.1 & 9.6 \\ \hline
    \end{tabular}
\end{table*}

\subsection{TPU results}
\label{app:tpu}

In this section, we report FNet efficiency results for TPUs; the main text focuses on GPUs. Figure \ref{fig:tpu_speed_accuracy} shows the speed vs MLM pre-training accuracy curve when training on TPU ($4\times4$ v3 chips). As on GPUs, FNet and the Linear model define the Pareto efficiency frontier for smaller, faster models, while BERT defines the frontier for larger, slower models.

Table \ref{tab:lra_training_speed_tpu} shows Long Range Arena Text classification efficiency results on TPUs ($4\times4$ v3 chips). The Linear model and FNet train faster than all the efficient Transformers for sequence lengths $\leq2048$ and $512$, respectively. For longer sequences, FNet is slower than the Performer and, based on results in \citet{tay2020long}, likely also slower than the other efficient Transformers that linearize attention, namely Local Attention \citep{parmar2018image}, Linformer \citep{wang2020linformer} and Linear Transformer \citep{katharopoulos2020transformers}. However, it is worth noting that Table \ref{tab:lra_accuracy} suggests that FNet is more accurate than all of the aforementioned models. Moreover, we expect that the GPU speed gains will transfer to TPUs as the TPU FFT implementation improves.

\begin{table*}
    \caption{Training (forward and backward passes; left) and inference (forward pass; left) speeds for \emph{only} the mixing sublayers -- all other model sublayers are removed. Both speeds are measured in milliseconds per batch (smaller is better), with batch sizes of 64 (GPU) and 256 (TPU). All batch examples have the sequence length fixed at 512. FNet uses the FFT for GPUs and matrix multiplications for TPUs. Speed up multipliers relative to self-attention are given in parentheses.}
    \label{tab:mixing_speeds}
    \centering
    \begin{tabular}{l | c c | c c}
        \hline
         & \multicolumn{2}{c|}{Training speed (ms/batch)} & \multicolumn{2}{c}{Inference speed (ms/batch)} \\ 
         & GPU & TPU & GPU & TPU\\ \hline \hline
         Self-attention (Base) & 136 & 76 & 43 & 16 \\
         Linear (Base) & 36 (3.7x) & 12 (6.1x) & 15 (2.8x) & \textbf{4 (3.9x)} \\
         FNet (Base) & \textbf{11 (12.2x)} & \textbf{8 (9.9x)} & \textbf{11 (4.0x)} & 8 (2.1x) \\ \hline
         Self-attention (Large) & 404 & 212 & 128 & 43 \\
         Linear (Large) & 103 (3.9x) & 35 (6.1x) & 36 (3.6x) & \textbf{10 (4.5x)} \\
         FNet (Large) & \textbf{18 (22.2x)} & \textbf{22 (9.7x)} & \textbf{18 (7.3x)} & 22 (2.0x)\\ \hline
    \end{tabular}
\end{table*}

\subsection{Additional configurations that we experimented with}
\label{app:things_we_tried}

We experimented with a number of additional ideas to improve FNet.

\textbf{Fourier Transform algorithm.} On GPUs, the FFT was the fastest algorithm for computing the DFT across all sequence lengths that we experimented with ($512-8192$). On TPUs, it is faster to compute the DFT directly using matrix multiplications for relatively shorter sequence lengths (up to lengths of $4096$; see Table \ref{tab:lra_training_speed_tpu}). This efficiency boundary between matrix multiplication and FFT on TPUs will change depending on the XLA precision for the matrix multiplications. We found that, although (slower) HIGHEST XLA precision was required to very accurately reproduce FFT in computing the DFT, (faster) DEFAULT XLA precision was sufficient to facilitate accurate model convergence.

\textbf{Modifying the Fourier Transform computation.} To keep the entire FNet architecture simple, the Fourier sublayer accepts real input and returns real output. The standard Fourier sublayer in FNet simply extracts the real part after computing the 2D DFT. We found that FNet was less accurate and less stable during training if only the real part of the DFT was used throughout the computation. Simply extracting the absolute value (instead of the real part) also led to a significantly less accurate model. Because the feed-forward sublayer mixes the hidden dimension, we experimented with  applying a 1D DFT along the token dimension only in the Fourier sublayer (i.e. no hidden dimension mixing in the Fourier sublayer). This yielded some training speed gains but hurt accuracy. The 1D (token mixing only) DFT model still significantly outperformed the (no token mixing) FF-only model, indicating that token mixing is most important mechanism in the Fourier sublayer.

\textbf{Other transforms.} We experimented with three natural alternatives to the Fourier Transform:
\begin{itemize}
    \item Discrete Cosine Transform (DCT). The DCT is closely related to the DFT but transforms real input to real output. However, we found that the DCT model underperformed FNet ($\sim4\%$ accuracy degradation).
    \item Hadamard Transform\footnote{Whereas the DFT matrix in Equation \eqref{eq:vandermonde} contains $N$ roots of unity, the Hadamard Transform simply contains two roots of unity: $\{\pm1\}$; see also \citet{kunz1979equivalence}.}. Although the Hadamard Transform was slightly faster than the DFT, it yielded less accurate results ($\sim2\%$ accuracy degradation).
    \item Hartley Transform. The Hartley Transform, which transforms real input to real output, can be described in terms of the Fourier Transform: $\mathcal{H} = \Re\left\{\mathcal{F}\right\} - \Im\left\{\mathcal{F}\right\}$. We found that the Hartley Transform matched the Fourier Transform on GLUE ($76.7$ vs. $76.7$). 
\end{itemize}

\textbf{Introducing learnable parameters to the Fourier sublayer.} Our attempts to introduce learnable parameters into the Fourier sublayer were either detrimental or inconsequential, and generally slightly slowed the model. For the (sequence length, hidden dimension) input in each Fourier sublayer, we tried two approaches to introduce learnable parameters: (1) element wise multiplication with a (sequence length, hidden dimension) matrix, and (2) regular matrix multiplication with (sequence length, sequence length) and (hidden dimension, hidden dimension) matrices. We experimented with these approaches in various configurations: preceding and/or following the DFT, and also in combination with inverse DFT (e.g. transform to frequency domain, apply element wise multiplication, transform back to time domain), but most setups degraded accuracy and reduced training stability, while a few did not change accuracy but lead to small speed decreases. In a slightly different set of experiments and in an effort to provide more flexibility to the model, we added (complex) learnable weights to the 2D DFT matrix. This model was stable but did not yield any accuracy gains, suggesting that the DFT is locally optimal in some sense.

\textbf{FNet block modifications.} The standard FNet encoder block structure follows that of the Transformer: a Fourier sublayer followed by a feed-forward sublayer, with residual connections and layer norms after each sublayer; see Figure \ref{fig:architecture}. We tried several modifications to this structure, based on the intuition of moving in and out of the frequency domain between multiplications. For example, the sandwiching of Fourier, feed-forward, Fourier (or inverse Fourier) sublayers and only applying the residual connections and layer norms to the final result, yields a structure that more closely mimics convolutions. However, these setups degraded accuracy and lead to a more unstable model during training. Adding extra feed-forward sublayers to this layering, or swapping out the feed-forward sublayers for simpler dense sublayers, did not help either.


\subsection{Mixing layer speeds}
\label{app:mixing_speeds}

Table \ref{tab:mixing_speeds} summarizes the inference and training speeds for the different mixing layers. For each of the Base and Large configurations, we have removed all other sublayers and transformations and then calculated the speed per batch of input examples. The FNet training speeds are particularly fast because no parameters are updated. The Linear model has faster inference than FNet on TPUs because it is performing real matrix multiplications, whereas FNet performs complex matrix multiplications; see Equation \eqref{eq:vandermonde}.

Although the Fourier mixing sublayer itself performs forward and backward passes significantly faster than the self-attention sublayer, FNet is overall 70-80\% faster than BERT because the overall training and inference speeds are bottle-necked by the feed-forward sublayers that all models share.

\subsection{FNet-Hybrid ablations}
\label{app:hybrid_ablations}

\begin{table}
    \caption{GPU pre-training accuracy and speed ablations for FNet-Hybrid models in the Base configuration. Batch size is 64. Metrics are recorded after 100k steps, which we have generally found to be a good indicator of final relative performance. See text for a description of the layouts.}
    \label{tab:hybrid_ablations}
    \setlength{\tabcolsep}{4pt}
    \centering
    \begin{tabular}{c c | c c | c}
        \hline
        \multicolumn{2}{c|}{Attention} & \multicolumn{2}{c|}{Accuracy} & Speed \\
        Layers & Layout & MLM & NSP & (ms/batch) \\ \hline \hline
        2 & BOTTOM & 0.497 & 0.733 & \textbf{193} \\
        2 & MIDDLE & 0.499 & 0.686 & 196 \\
        2 & MIXED & 0.509 &	0.727 &	194 \\
        2 & TOP & \textbf{0.526} &	\textbf{0.738} &	\textbf{193} \\ \hline
        0 & TOP & 0.486 &	0.679 &	\textbf{173} \\
        2 & TOP & 0.526 &	0.738 &	193 \\
        4 & TOP & 0.539 &	0.740 &	214 \\
        6 & TOP & \textbf{0.546} & \textbf{0.746}	& 235 \\ \hline
    \end{tabular}
\end{table}

Table \ref{tab:hybrid_ablations} shows the effects of varying the number of attention sublayers and the attention layout in the FNet-Hybrid model. For the ``BOTTOM'' layout, all attention sublayers are placed in the first few encoder layers, where they replace the Fourier mixing sublayers. For the ``TOP'' layout, attention sublayers are placed in the final encoder layers; for the ``MIDDLE'' layout they are placed in the middle layers; and for the ``MIXED'' layout, they are distributed through the model.

From the Table \ref{tab:hybrid_ablations}, we can make two observations: (1) more attention improves accuracy at the cost of speed, and ultimately with diminishing returns; (2) placing attention layers at the top of the model gives the best accuracy results. Given our focus on speed, we chose to focus FNet-Hybrid experiments in the main text of the paper on the 2 attention layer, ``TOP'' configuration variant.

\subsection{A note on Long-Range Arena hyperparameter settings}
\label{app:lra}

Concerning the Long-Range Arena setup, several hyperparameters are not described in \citet{tay2020long} and there a few mismatches between the configurations described in the paper and the code repository. Where possible, we prioritize configurations described in the paper with only two exceptions. Firstly, for the CIFAR10 (Image) task, we perform a sweep of the number of layers in the range $[1, 2, 3, 4]$. We found that 1 layer worked best for all models; \citet{tay2020long} suggest 3 layers yielded the best results. Secondly, for the Pathfinder task, we found that a base learning rate of $0.001$ (as given in the code repository) yielded better results for all models than the $0.01$ value indicated in \citet{tay2020long}.  We also perform a very small sweep over the embedding dimension and batch size, which are not listed in \citet{tay2020long}.

We also remark that the accuracy comparisons between our runs and those from \citet{tay2020long} should be performed with the caveat that we found that results for certain tasks -- Text and Retrieval in particular -- can vary quite a bit between runs, especially for the Transformer; we report the best results.

\subsection{FNet code}
\label{app:code}

\begin{lstlisting}[language=python, caption=FNet code written in JAX/Flax. Embedding and output projection layers are omitted for simplicity., captionpos=b, float=*t]
import flax.linen as nn
import jax
import jax.numpy as jnp


class FourierTransformLayer(nn.Module):
  @nn.compact
  def __call__(self, x):
    return jax.vmap(jnp.fft.fftn)(x).real
    
    
class FeedForwardLayer(nn.Module):
  d_ff: int
  dropout_rate: float
  
  @nn.compact
  def __call__(self, x, deterministic):
    x = nn.Dense(self.d_ff, 
      kernel_init=nn.initializers.normal(2e-2), 
      bias_init=nn.initializers.normal(2e-2), 
      name="intermediate")(x)
    x = nn.gelu(x)
    x = nn.Dense(x.shape[-1], 
      kernel_init=nn.initializers.normal(2e-2), 
      name="output")(x)
    return nn.Dropout(self.dropout_rate)(x, deterministic)


class FNetEncoderBlock(nn.Module):
  fourier_layer: FourierTransformLayer
  ff_layer: FeedForwardLayer
  
  @nn.compact
  def __call__(self, x, deterministic):
    mixing_output = self.fourier_layer(x)
    x = nn.LayerNorm(1e-12, name="mixing_layer_norm")(x + mixing_output)
    feed_forward_output = self.ff_layer(x, deterministic)
    return nn.LayerNorm(
        1e-12, name="output_layer_norm")(x + feed_forward_output)
        
        
class FNetEncoder(nn.Module):
  num_layers: int
  d_model: int
  d_ff: int
  dropout_rate: float
  
  def setup(self):
    encoder_blocks = []
    for layer in range(self.num_layers):
      encoder_blocks.append(FNetEncoderBlock(
        FourierTransformerLayer(), 
        FeedForwardLayer(self.d_ff, self.dropout_rate), 
        name=f"encoder_{layer}"))
    self.encoder_blocks = encoder_blocks
    self.pooler = nn.Dense(
        self.d_model, 
        kernel_init=nn.initializers.normal(2e-2), 
        name="pooler")

  def __call__(self, x, deterministic):
    for encoder_block in self.encoder_blocks:
      x = encoder_block(x, deterministic)
    pooled_output = self.pooler(x[:, 0])
    pooled_output = jnp.tanh(pooled_output)
    return x, pooled_output
\end{lstlisting}

\end{document}